\documentclass[letterpaper, 10 pt, conference]{ieeeconf}  
\usepackage[utf8]{inputenc}
\usepackage{amsmath, amsfonts}
\usepackage{graphicx}
\usepackage{xcolor}
\usepackage{wrapfig}
\usepackage{adjustbox}
\usepackage{xspace}
\usepackage{epstopdf}
\usepackage{xcolor}
\usepackage{float}
\usepackage{soul}
\usepackage{algorithm}
\usepackage{algcompatible}
\usepackage[capitalize]{cleveref}

\usepackage{eqexpl}
\eqexplSetDelim{=}

\usepackage{algpseudocode}
\usepackage{array}
\usepackage{tabularx}

\usepackage{wrapfig}
\usepackage[small]{caption}
\usepackage{subcaption}
\usepackage[noadjust]{cite}
\usepackage{comment}

\IEEEoverridecommandlockouts 

\newcommand{\systemname}{\textit{}\xspace}

\title{\systemname A Sensor Position Localization Method\\ for Flexible, Non-Uniform Capacitive Tactile Sensor Arrays}

\author{Carson Kohlbrenner, Caleb Escobedo, Nataliya Nechyporenko, and Alessandro Roncone 
\thanks{ All authors are with the Department of Computer Science, University of Colorado Boulder, 1111 Engineering Drive, Boulder, CO USA. 
Nataliya is supported by NSF DGE \#2040434. This work is partially supported by NSF FW-HTF-R \#2222952.
{\tt\small name.surname@colorado.edu}.}
}
\begin{document}
\maketitle

\begin{abstract}
Tactile sensing is used in robotics to obtain real-time feedback during physical interactions. Fine object manipulation is a robotic application that benefits from a high density of sensors to accurately estimate object pose, whereas a low sensing resolution is sufficient for collision detection.
Introducing variable sensing resolution into a single tactile sensing array can increase the range of tactile use cases, but also invokes challenges in localizing internal sensor positions.
In this work, we present a mutual capacitance sensor array with variable sensor density, \textsl{VARSkin}, along with a localization method that determines the position of each sensor in the non-uniform array. When tested on two distinct artificial skin patches with concealed sensor layouts, our method achieves a localization accuracy within $\pm 2$mm. 
We also provide a comprehensive error analysis, offering strategies for further precision improvement.
\end{abstract}
\section{Introduction}

\begin{figure}
    \vspace{0.8em}
  \centering
  \includegraphics[width=\linewidth]{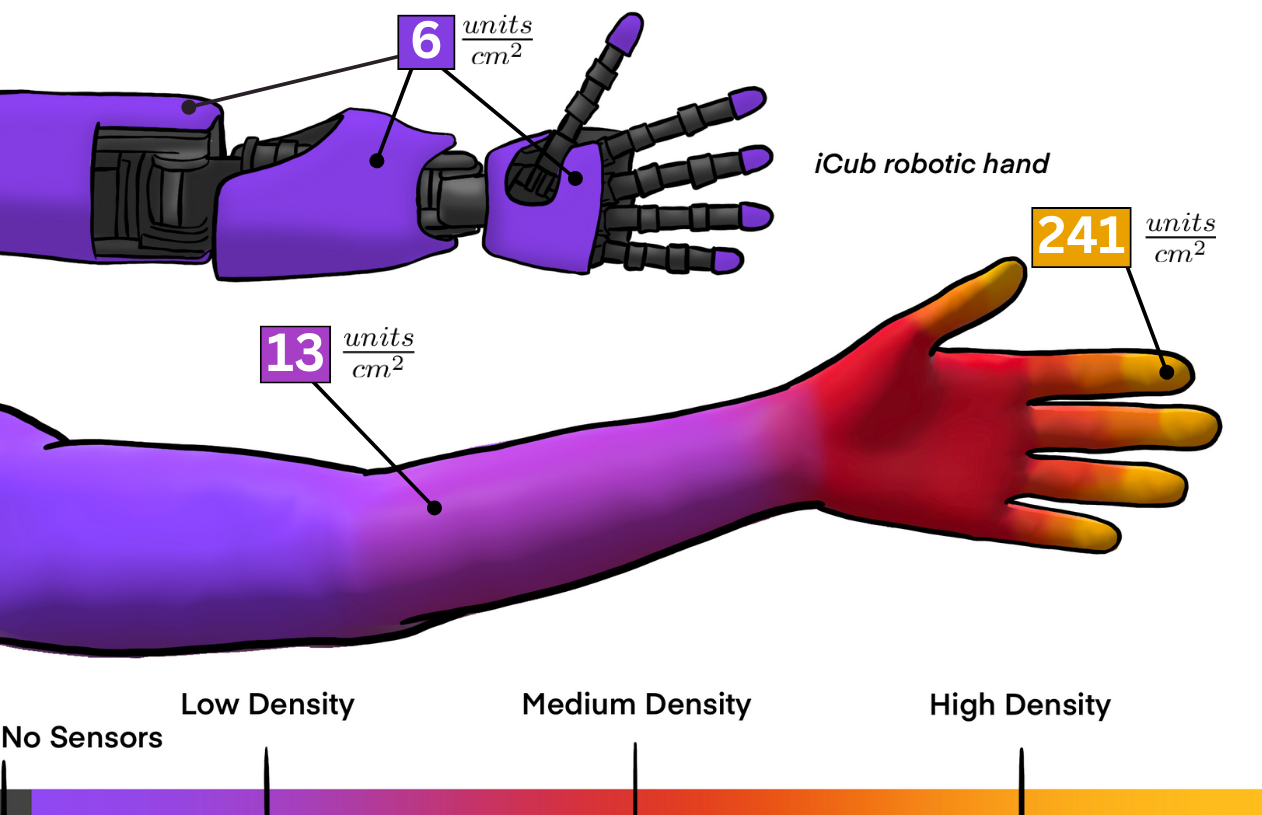}
  \caption{The prevalent large-scale    
    artificial skin implementation in the iCub robot has equal average sensor density across the robot arm. 
    In contrast, the average
    biological sensor densities of human skin gradually increase from the base of the arm to the tip of the fingers. 
    In our work, we aim to mimic the biological approach of variable sensor density which we see in humans.
    \cite{corniani2020tactile, maiolino2013flexible}.}
  \vspace{-1.5em}
  \label{fig:sensorDensity}
\end{figure}

Artificial skin technologies, which generally aim at replicating the tactile sensing attributes of humans', find applications in diverse domains such as healthcare, prosthetics, and robotics \cite{Yang2019}. 
While numerous frameworks have been proposed, a conspicuous gap remains: the lack of a scalable, variable-resolution architecture suitable for both \textsl{large scale} (e.g., to cover all or the majority of a robot's body) and \textsl{small scale} (e.g. to perform fine manipulation) implementations. 
Existing high-fidelity sensor technologies such as \textsl{DenseTact} and \textsl{GelSight} excel in emulating the tactile resolution of human fingertips but are unable to cover large surface areas \cite{do2023densetact, do2022densetact, yuan2017gelsight}.
Conversely, designs aimed at large surface coverage tend to compromise on tactile resolution \cite{teyssier2021human, cheng2019comprehensive, maiolino2013flexible, schmitz2011methods}.
This dichotomy overlooks the innate versatility of human skin, as seen in Fig. \ref{fig:sensorDensity}, which exhibits variable sensor density tuned to the functional requirements of different anatomical regions \cite{corniani2020tactile, dvorak2021ultrathin}.
It is precisely this adaptive feature that existing artificial skins, characterized predominantly by their uniformly-distributed sensor arrays, fail to capture \cite{Dahiya2019Humanoids, Dahiya2013, cheng2019comprehensive, teyssier2021human, navarro2021proximity, bergner2020design}.
Much like the nuanced tactile resolution across different areas of the human body, the integration of artificial skin into robotic systems calls for a tailored approach to sensor density, particularly when these systems are designed for multifaceted applications ranging from locomotion to fine manipulation.

Engineering artificial skins with optimized sensor distribution offers tangible benefits for robotic applications, including enhanced efficiency in data acquisition, power consumption, and wiring architecture, primarily through the strategic elimination of extraneous sensors \cite{Dahiya2019LargeArea}.
Despite the benefits, a significant challenge emerges: the absence of a straightforward method to determine the relative locations of individual sensors. This issue becomes accentuated when the skin either conceals sensor locations or employs non-standard manufacturing processes, such as randomized sensor arrangements. Without known sensor locations,  robots struggle to perform tasks such as object orientation detection and manipulation \cite{erickson2019multidimensional, dong2017improved}.
 
\begin{figure}
  \centering
  \includegraphics[width=\linewidth]{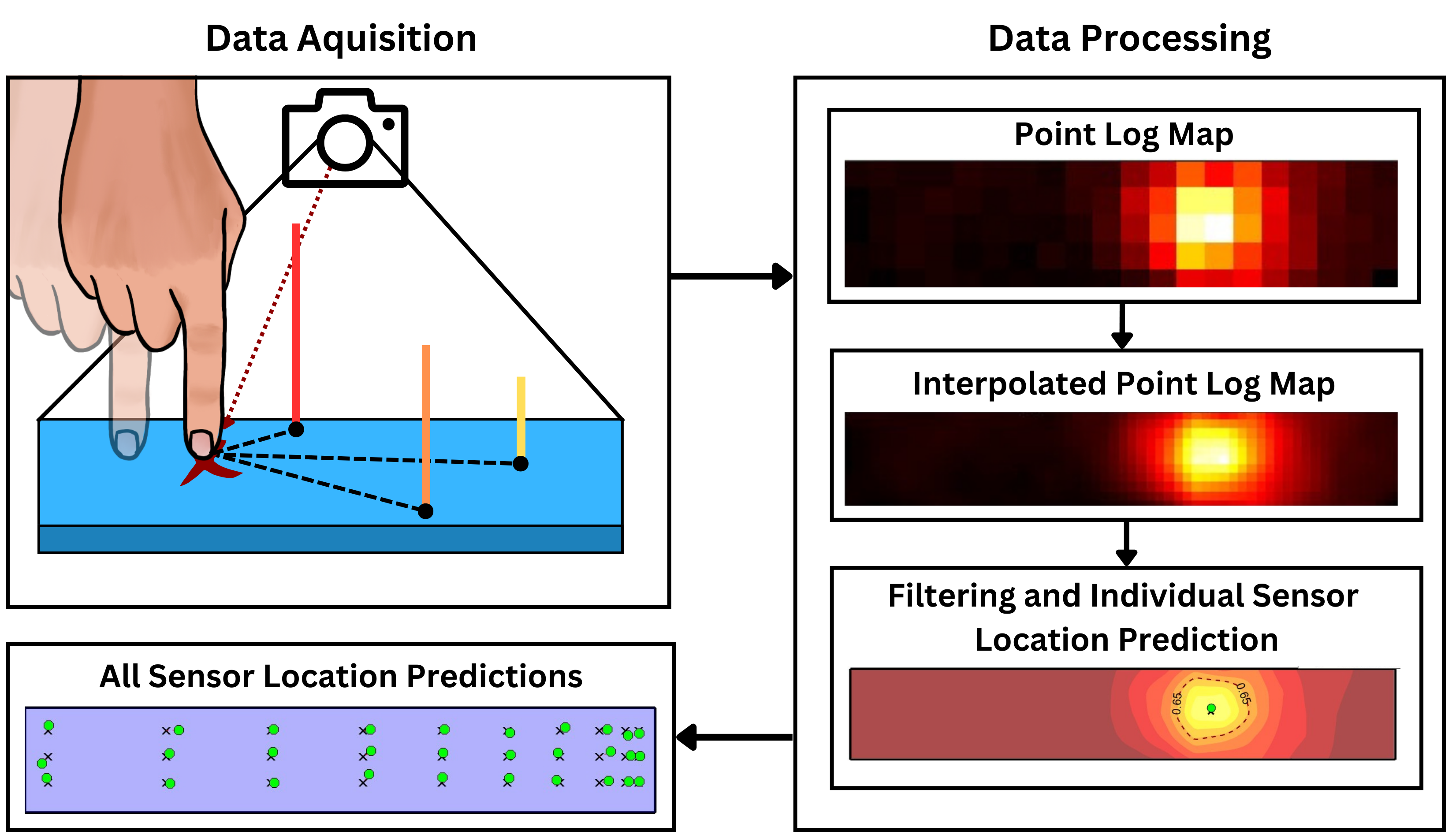}
  \caption{In this work, we propose a localization method to predict the location of each sensor in a non-uniform sensor density artificial skin. The method starts by constructing a \textsl{point log map} that shows a single sensor's recorded intensities at different probing locations. The \textsl{point log map} is then interpolated, filtered, and averaged, resulting in a single sensor location prediction. This method is then repeated for each sensor to compile the final sensor location predictions.}
  \vspace{-1.5em}
  \label{fig:diagram}
\end{figure}
In this paper, we introduce what we believe is the first example of an artificial skin with non-uniform sensor density.
This is complemented by a computational method to determine the location of each tactile sensor inside the skin, which we validate on real hardware with two distinct artificial skin patches. 

Our contributions are as follows:
a) we present a \textsl{flexible, variable-density artificial skin} equipped with mutual capacitance sensors. The skin's variable sensor distribution is fully compatible with our novel localization method, ensuring robust performance;
b) we propose a systematic, repeatable \textsl{localization method} designed explicitly for mutual capacitance sensors. This algorithm accurately localizes each embedded sensor, making the use of this skin technology possible for applications requiring detailed spatial information;
c) we validate the efficacy of our algorithm by applying it to two distinct artificial skin samples with concealed sensor layouts. Further, we conduct an extensive error analysis to identify and address potential avenues for optimization.
Collectively, the contributions presented in this work not only advance the field of tactile sensing but also pave the way for more versatile and adaptive robotic systems. By tailoring sensor density and offering precise localization, our artificial skin can be integrated into a large array of robotic applications requiring complex spatial awareness, from fine object manipulation and perception to dynamic locomotion.

\section{Related Work}
   
\subsection{Past and Emerging Trends in Artificial Skins for Robotics}
  In recent years, many artificial skin technologies have emerged that can measure force, strain, slip, heat, and other sensations related to the sense of touch.
  High resolution artificial skins have been proposed with acoustics \cite{wall2023passive, rupavatharam2023sonicfinger}, computer vision \cite{chen2018tactile, do2022densetact, do2023densetact, dong2017improved}, capacitive proximity sensing \cite{teyssier2021human, ding2019proximity, goeger2010tactile, goger2013tactile, hoshi2006robot, navarro2013methods, wang2013mutual, maiolino2013flexible}, and dense sensor packages \cite{mittendorfer2011humanoid}. 
  Of note, large-scale artificial skins have been developed and successfully deployed on platforms such as the iCub and REEM-C robots, albeit with unoptimized, uniform sensor densities \cite{maiolino2013flexible, cheng2019comprehensive}. 
  The artificial skin architecture proposed in this work was inspired by the mutual capacitance sensing paradigm articulated by \cite{teyssier2021human}, and adopts and builds upon the Muca development board to record sensor data. 

  Beyond the advancements in artificial skin technology, there is a corresponding need for solutions that maximize the utility of large-scale robotic skins, particularly in the domain of machine learning-based robotics \cite{Dahiya2019LargeArea}. 
  Varied sensor density in artificial skins emerges as a pivotal factor, potentially reducing computational overhead by selectively emphasizing data-rich or mission-critical regions on the robot.
  This optimization not only minimizes computational and energy costs but also enriches the quality of feedback data, a crucial component in the success of reinforcement learning algorithms \cite{bhagat2019deep, amarjyoti2017deep}.
  Other avenues of efficiency beyond optimized density distribution include reducing the overall number of sensors, a factor which is highly dependent on both the geometry and the specific application of the tactile sensor arrays \cite{massari2020machine}.

\subsection{Algorithms for Kinematic Calibration of Robotic Skin with Concealed Sensor Layouts}
    \cite{watanabe2021self} presents a kinematic approach to determining the location of sensors in artificial skin. This approach requirses an Inertial Measurement Unit (IMU) device for each sensor to calculate relative positions, however, IMUs lack flexibility and can be expensive to scale down if high-resolution artificial skin is necessary. Another kinematic approach demonstrated by \cite{roncone2014automatic} uses self-touch for position calibration via artificial skin built into the iCub humanoid robot. This approach kinematically localizes the positions of the sensors using joint configuration calibration and the relative positions of the sensors on the joints. Our approach does not require any additional components unnecessary to the artificial skin's design or knowledge of joint poses to determine sensor locations.

\section{Methods}


 \begin{figure}[b!]
  \centering
  \includegraphics[width=1\linewidth]{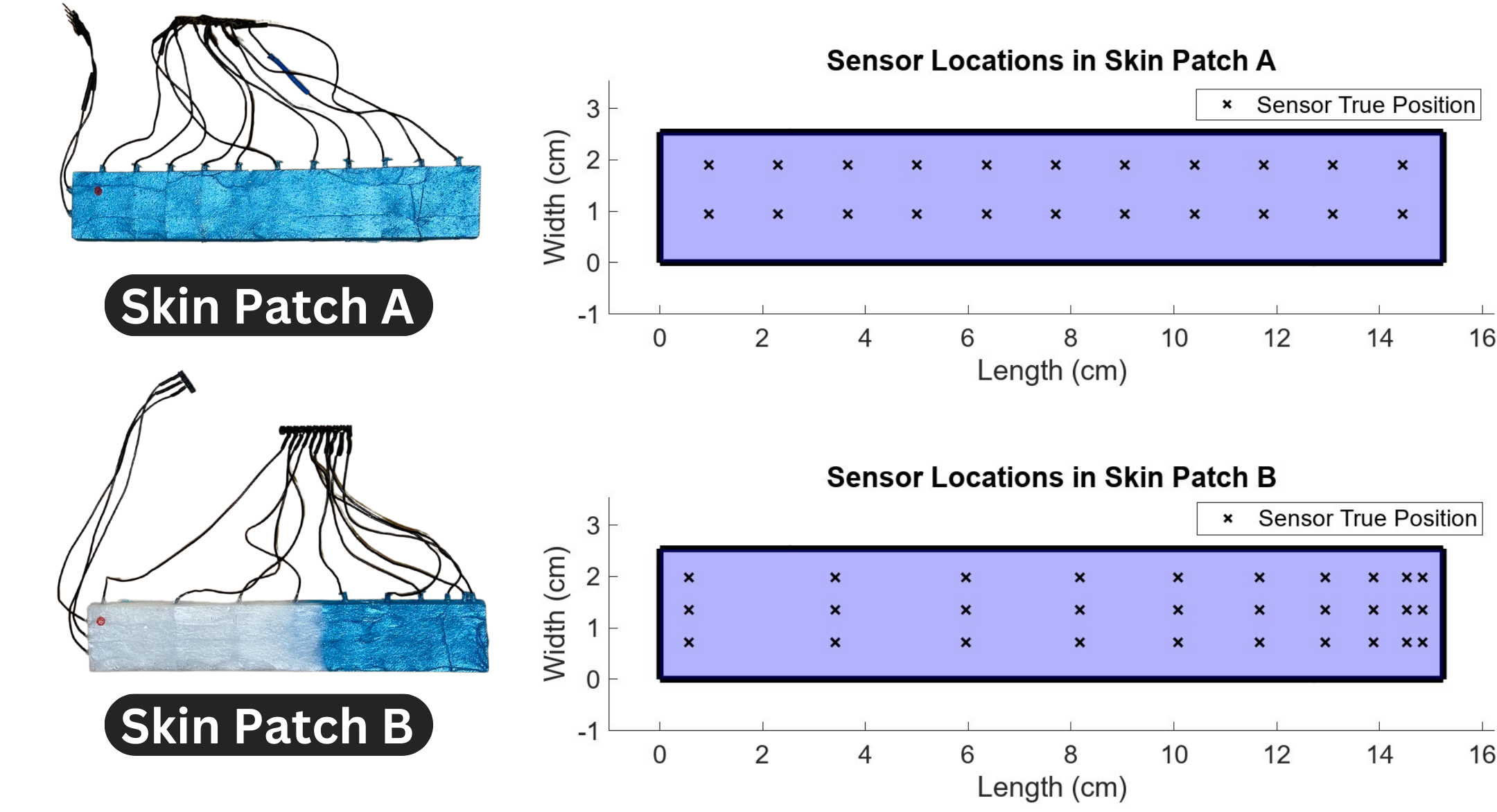}
  \caption{We fabricated two $2.54$x$15.24cm$ skin sensor patches capable of real-time tactile perception. Top blue sensor - \textsl{Patch A} - has an even distribution of internal electrodes which is a predominant feature in state-of-the-art approaches. The bottom white and blue sensor - \textsl{Patch B} - has a variation of electrode locations along its length and demonstrates our novel approach to enable variable-density skin. Our contributed localization method generalizes to \textsl{both} configurations of electrodes.}
  \label{fig:skinsensors}
\end{figure}

\subsection{Design of a Variable-Density Robotic Skin: VARSkin}\label{sec:system_specs}

This section outlines the methods and materials utilized to fabricate our proposed variable-density skin, which we refer to as \textsl{VARSkin}. The primary aim is to produce a flexible artificial skin capable of delivering high-quality tactile data for diverse robotic applications, ranging from simple touch detection to complex human-robot interactions.
Two $2.54$x$15.24cm$ patches were fabricated using \textsl{Smooth-On Dragon Skin 10} silicone rubber to emulate the flexibility and elasticity of human skin. This material also functions as the insulator layer to achieve effective capacitive sensing. A 3D-printed PLA mold featuring cutouts along the edges was used for silicone casting, enabling the threading of a silicone-covered stranded-core wire before curing. 
With reference to \cref{fig:skinsensors}, the first skin patch (referred to in this paper as \textsl{Patch A}) consists of eleven equidistant cutouts spaced $1.35 cm$ apart longitudinally and two additional cutouts spaced $1 cm$ apart along its width; 
the second skin patch (i.e. \textsl{Patch B}) has ten cutouts spaced irregularly along its length, ranging from $2.86cm$ to $0.32cm$, and three cutouts $0.63cm$ apart along its width.
After curing, a flexible copper sheet is adhered to the bottom of each patch.
The wires are then soldered to header pins and connected to a \textsl{Muca} capacitive sensing development board \cite{teyssier2021human}. 
When grounded, the copper serves as an electromagnetic shield, isolating the sensor measurements to only capture data from the exposed skin surface, thereby minimizing external noise.

\subsection{Sensor Placement and Function}

\begin{figure} \vspace{1.5em}
    \centering
    \subfloat[\centering]{{\includegraphics[width=0.21\textwidth]{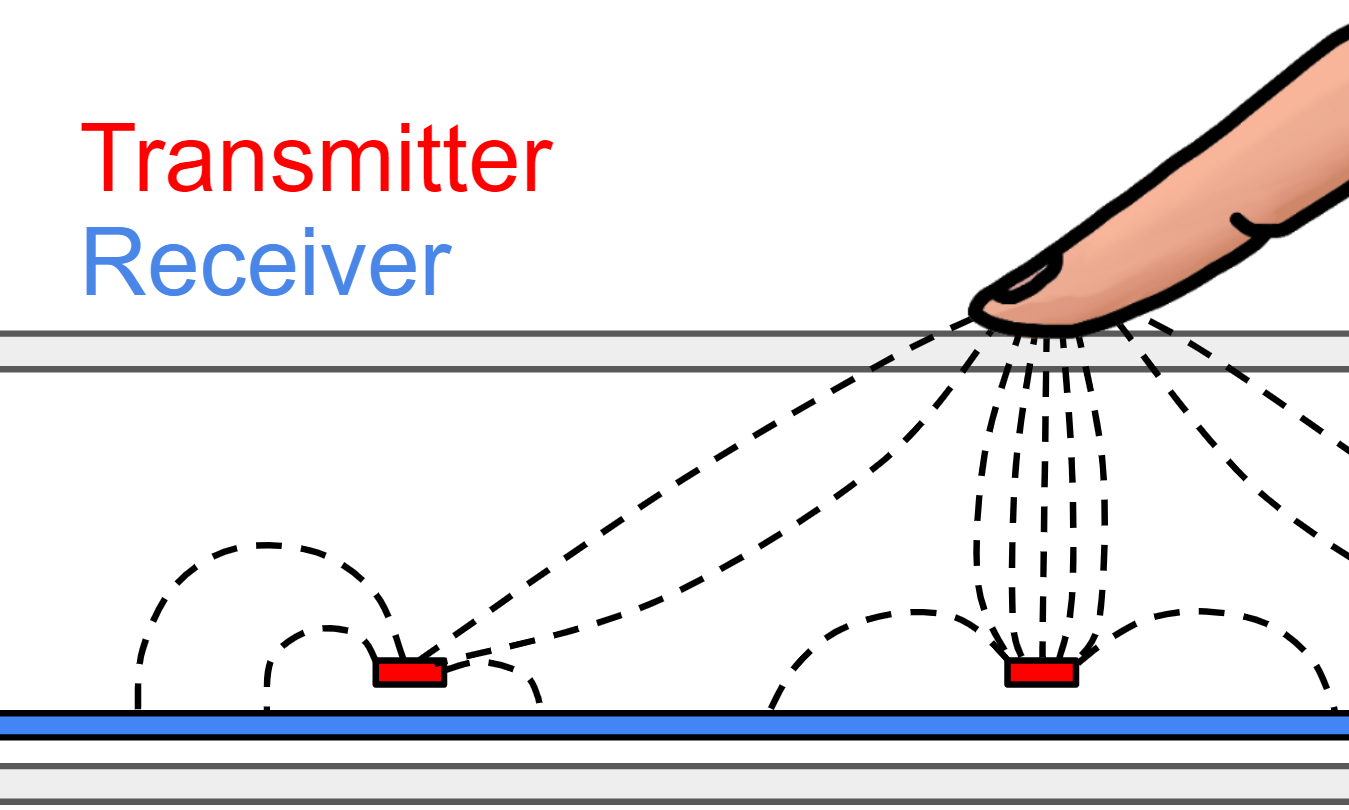} }}%
    \qquad
    \subfloat[\centering]{{\includegraphics[width=0.21\textwidth]{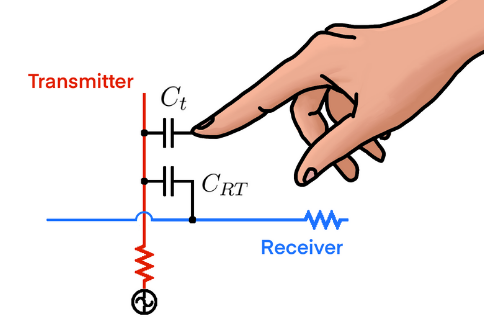} }}
    \caption{
    Our tactile skin patch detects touch by measuring changes in mutual capacitance, which is an electromagnetic field created by two overlapping electrodes. 
    a) The electromagnetic field when a human finger is in close proximity to an electrode crossing point. b) A circuit diagram of the \textsl{transmitter} and \textsl{receiver} electrodes forming a mutual capacitor $C_{RT}$ in parallel with a human finger in close proximity.}
    \label{fig:circuit}%
\end{figure}

  Mutual capacitance sensors can measure a change in the nearby electromagnetic field by creating a potential difference through a \textsl{transmitter} electrode that induces a cyclic charge pattern measured in a \textsl{receiver} electrode. The design of \textsl{VARSkin} incorporates two distinct electrode configurations. 
  \textsl{Patch A} features $11$ transmit and $2$ receive electrodes, creating $22$ evenly spaced intersection points. In contrast, \textsl{Patch B} is equipped with $10$ transmit and $3$ receive electrodes, yielding $30$ intersection points with variable spacing. 
  With reference to Fig. \ref{fig:circuit}, at each intersection of transmit and receive electrodes, a capacitor, denoted as $C_{RT}$, is formed. This capacitance functions as a proximity sensor and can be measured by the Muca board.
  When an external object---such as a human finger or a grounded probe---enters the electromagnetic field at the intersection, an additional capacitor $C_t$ forms in parallel with $C_{RT}$. This new capacitor \textsl{reduces} the amount of charge transferred to the receiver electrode, causing the Muca board to register a \textsl{higher} capacitance value for that particular sensor.

\begin{figure}\vspace{0.2em}
  \centering
  \includegraphics[width=\linewidth]{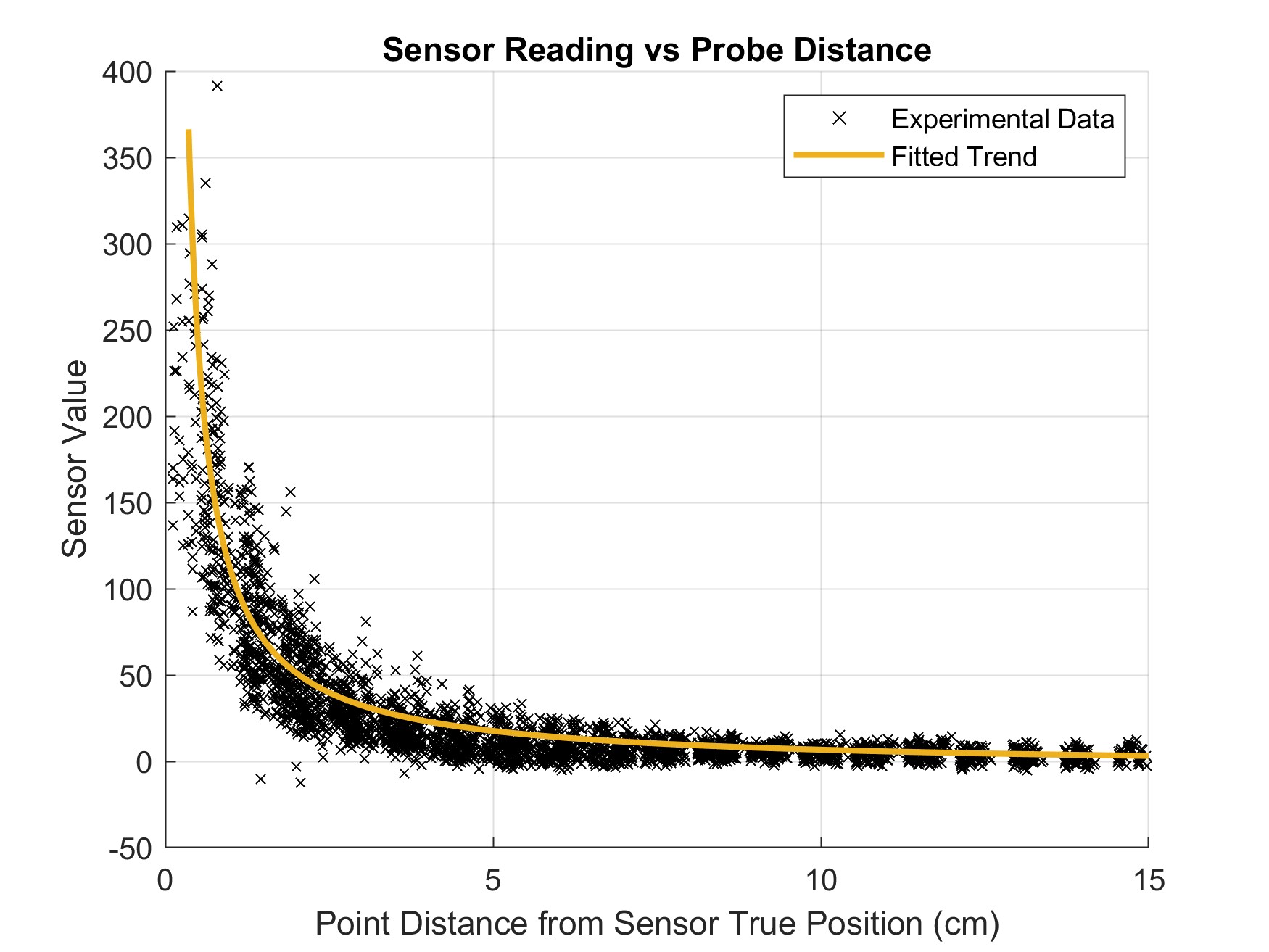}
  \caption{
  A probe was lightly pressed against a $5$x$20$ equally spaced grid of locations along the surface of the artificial skin.
  This figure shows the distances between the probe location and sensor locations plotted against the capacitance measurements of each sensor.
  The resulting trend is used in our novel localization method to predict the sensor locations in \textsl{Patch B} from a light touch.
  }
  \label{fig:sensordist}
\end{figure}

  To better understand the sensor's responsiveness, $100$ different locations on \textsl{Patch B} were probed and the correlated sensor values were recorded. The results, illustrated in Fig. \ref{fig:sensordist}, reveal an asymptotic increase in sensor output as the probe distance decreases. This relationship is fundamental to the prediction of sensor locations with the localization method presented.

\subsection{Localization Algorithm}

To predict sensor locations, we generate, interpolate, and average a topographic map (see \cref{fig:3dinterp}) generated from the sensors' measures capacitance values at a set of known probe locations, which we hereinafter refer to as \textsl{point logs}. These are specific sets of capacitance values corresponding to distinct probe placements on the \textsl{VARSkin}.
The inverse correlation between a probe's placement along the \textsl{VARSkin} surface and the probe's distance to a given sensor creates a distinct hill shape that indicates sensor location when enough \textsl{point logs} are taken at varying sensor distances. For consistency and without loss of fidelity, all \textsl{point logs} taken in this paper were probed via a human finger.

Each \textsl{point log} needs a known and accurate probe location on the surface of the \textsl{VARSkin} for the localization method to perform accurately. This was achieved by predefining an equispaced $5$x$20$ grid that aligned with the \textsl{VARSkin}. The grid was digitally overlayed on the skin using a webcam feed; other techniques such as predefined physical markers are possible.

 Data is processed in three steps to predict sensor locations. These steps, detailed below, were repeated for each sensor in each skin patch. 

\begin{figure}\vspace{1em}
  \centering
  \includegraphics[width=\linewidth]{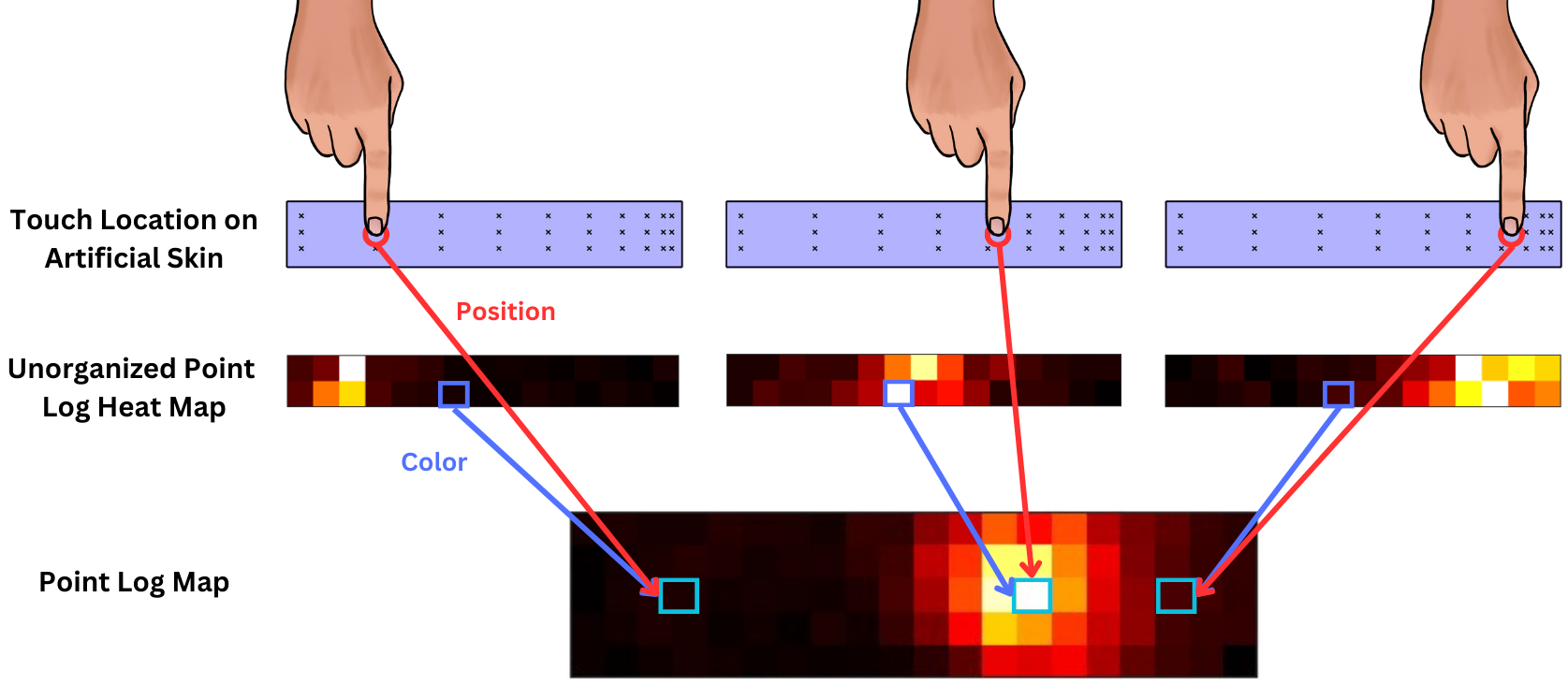}
  \caption{
   We define a \textsl{point log} as a signal measurement recorded by all the sensors from a probe touching a pre-determined location on the skin patch.
  A \textsl{point log map} shows a heatmap of signal values for a single sensor after a set of \textsl{point logs} were recorded. The \textsl{point log map} shown here is for sensor 14 in \textsl{Patch B}.}
  \vspace{-1.5em}
  \label{fig:pointmap}
\end{figure}

  \subsubsection{Point Log Map} A special heat map, termed \textsl{point log map}, is computed to represent capacitance values for each sensor in the array as a function of finger presses at different locations on the array. This map differs from a traditional heat map \cite{teyssier2021human, navarro2013methods} in that, with reference to \cref{fig:pointmap}, each square represents a location on the surface of the artificial skin where the touch occurred, and its intensity represents the measured capacitance of the sensor at that touch location.
  The construction of a \textsl{point log map} for an individual sensor reading three different \textsl{point logs} and how these individual logs fit into a \textsl{point log map} is shown in Fig \ref{fig:pointmap}.

  \subsubsection{Interpolation} The data in a \textsl{point log map} is then upscaled using a piecewise cubic interpolant. This increases resolution while maintaining spatial accuracy. 
    This interpolation method retains spatial accuracy by preserving the continuous second derivative boundary condition expected of the asymptotic capacitance measurements with distance. Using this method, the raw \textsl{point log map} seen in Fig. \ref{fig:method}a is up-scaled to a higher resolution of $8$ pixels$/cm^2$ as seen in Fig. \ref{fig:method}b. 
    This techique has the advantage of blurring \textsl{point log map} pixels with their nearby neighbors and thus reduces the impact any outliers may have on the final prediction.

\begin{figure}
\vspace{0.8em}
  \centering
  \includegraphics[width=0.6\linewidth]{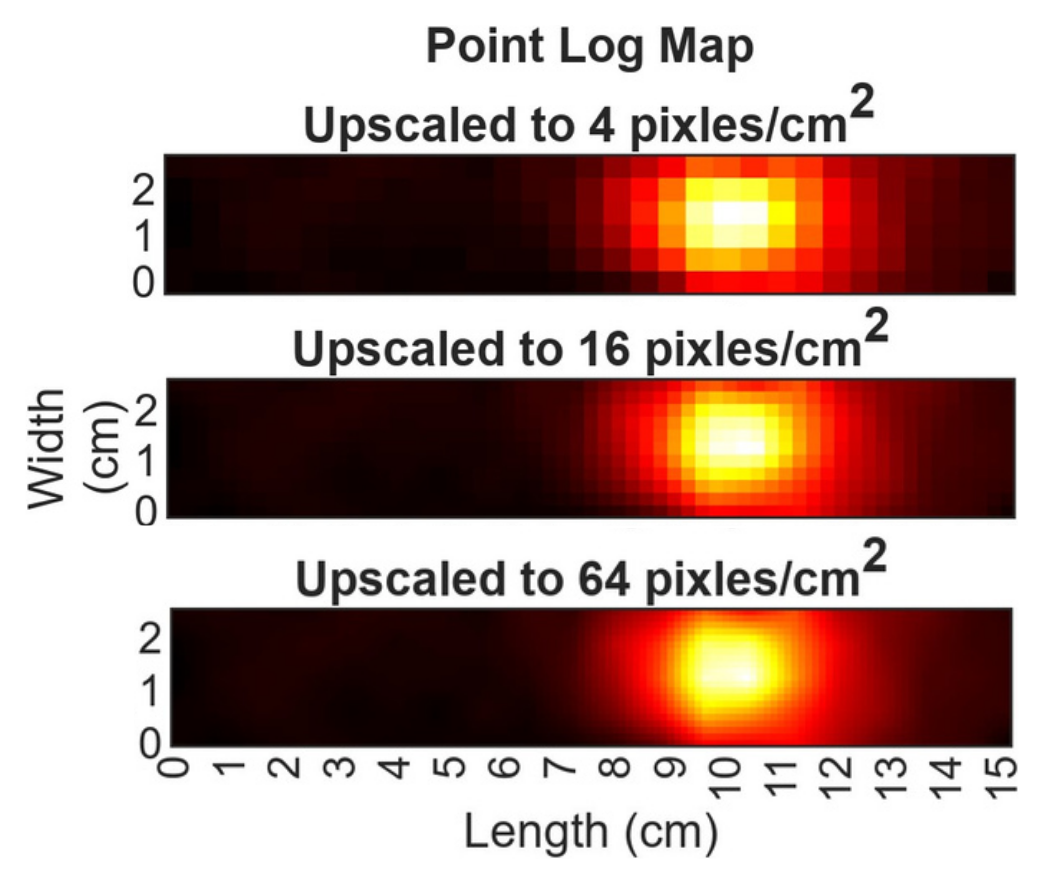}
  \caption{To increase the resolution while maintaining spatial accuracy during data processing, we perform an up-scaling of the the \textsl{point log map}. The following figure shows the up-scaling effect from Fig. \ref{fig:method}a at 4,16, and 64 $\text{pixels}/cm^2$.}

  \label{fig:ppcmCompare}
\end{figure}

  \subsubsection{Filtering} 

\begin{figure}\vspace{1.3em}
    \centering
    \subfloat[\centering]{{\includegraphics[width=0.22\textwidth]{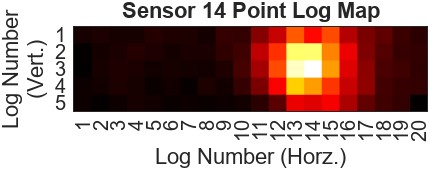} }}%
    \subfloat[\centering]{{\includegraphics[width=0.24\textwidth]{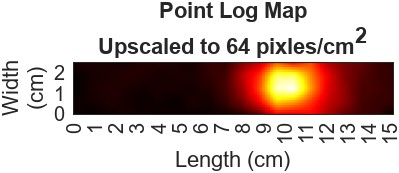} }}%
    \qquad
    \subfloat[\centering]{{\includegraphics[width=0.4\textwidth]{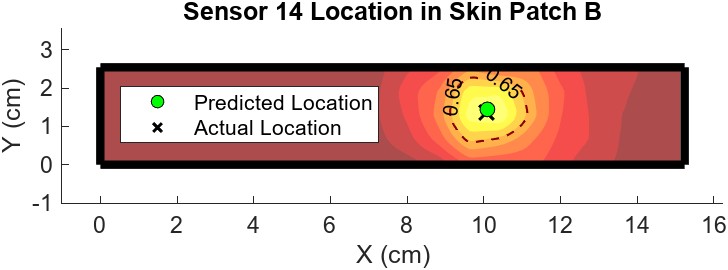} }}%
    \caption{
    The data collected by each sensor on the skin patch undergoes three processing steps. The following figure visualizes
    the processing results from sensor 14 in \textsl{Patch B}.
    a) In step one we generate a raw \textsl{point log map} when a 5x20 uniform grid of \textsl{point logs} are taken on the surface of the artificial skin. b) In step two the \textsl{point log map} is interpolated. c) In step three we apply filtering and averaging to generate the final predictions for the sensor.}%
    \label{fig:method}%
\end{figure}

\begin{figure}
  \centering
  \includegraphics[width=\linewidth]{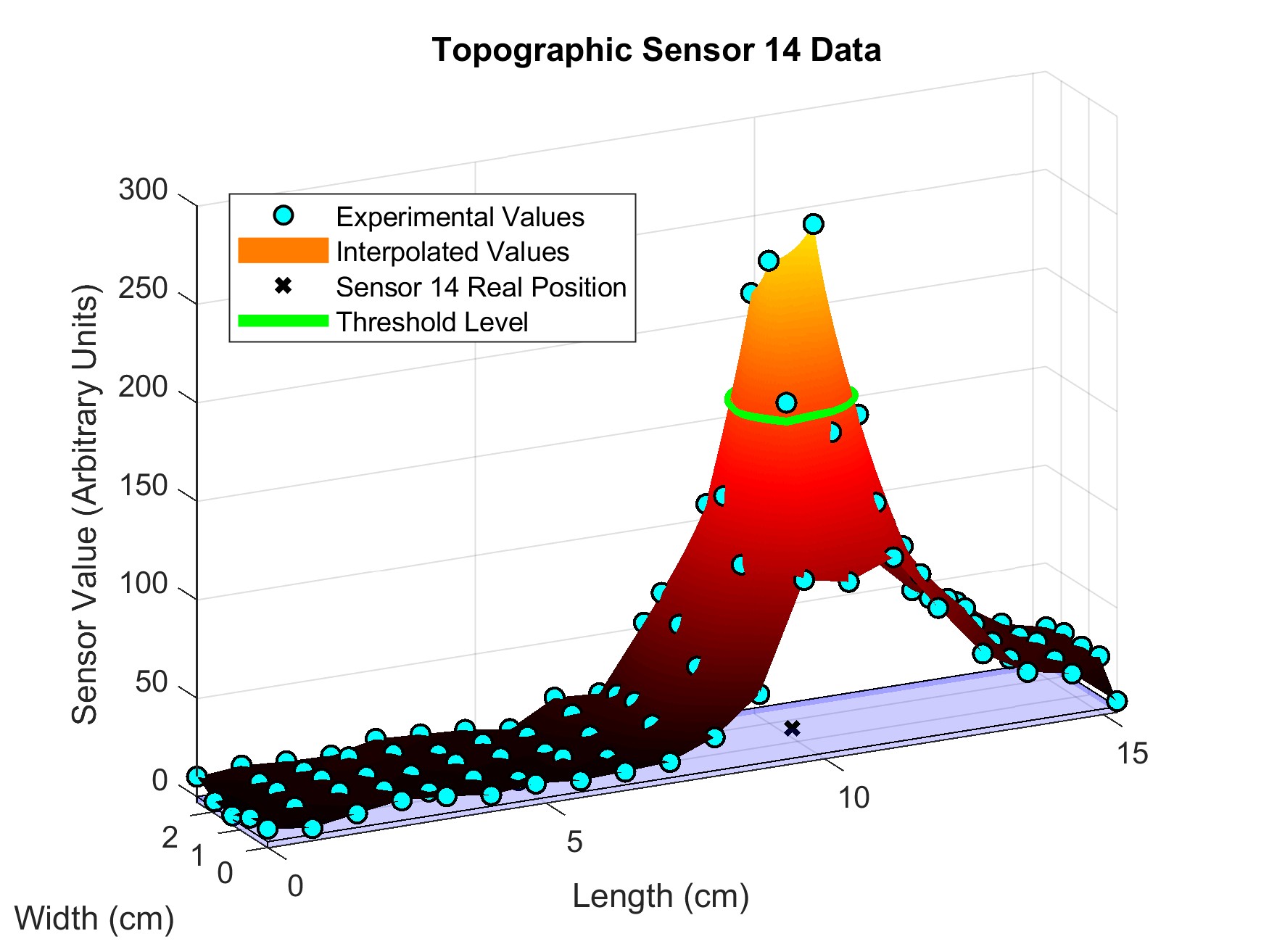}
  \caption{Topographic map of the interpolated \textsl{point log map} seen in \cref{fig:method}, where signal intensity is represented as elevation. The threshold value, seen in green, represents the cutoff for filtering. Every point above the threshold is considered in the final averaging for sensor location prediction.}
  \vspace{-1.5em}
  \label{fig:3dinterp}
\end{figure}

    The relationship between a sensor's output and its distance to a probe described in \cref{sec:system_specs} suggests that the expected shape of the interpolated \textsl{point log map} when plotted as a topographical map is a single spike centered over the sensor location. Fig \ref{fig:3dinterp} shows the topographic view of the \textsl{point log map} from \cref{fig:method}b along with the raw \textsl{point log map} data. 
    From this, the position of the spike is located by filtering out any data point under a specified percentage of the maximum recorded sensor value. An average position of all interpolated data points within the filtered spike is then calculated using Eq. \ref{eq:avgPos}:
    \begin{equation}
        (x,y)_{\text{pred}} = \frac{\Sigma (x,y)_{cc}}{|(x,y)_{cc}|}\qquad
        \label{eq:avgPos}
    \end{equation} 
    where:
    \begin{eqexpl}[14mm]
        \item{$(x,y)_{\text{pred}}$} Relative prediction coordinates from origin
        \item{$(x,y)_{cc}$} $\{(x,y)|I(x,y) > \eta \cdot \text{max}(I)\}$
        \item{$I$} Set of interpolated sensor values.
        \item{$\eta$} Threshold value.\\
    \end{eqexpl}
     The filtering process to get a sensor location prediction is demonstrated in \cref{fig:method}c with a threshold value of $\eta = 0.65$.

\section{Evaluation}
The accuracy of our non-uniform sensor localization method and the reliability of the artificial skin were evaluated using data collected upon full completion of the localization method with 100 \textsl{point logs} distributed in a 5x20 evenly spaced grid of locations on the surface of the \textsl{VARSkin}. The artificial skin was designed to return consistent capacitance measurements for sensor localization. The consistency of sensor measurements can be determined by their noise levels through a signal-to-noise ratio. A sample size of 50 sensor measurements was used for a no-contact calibration measurement and every \textsl{point log}. Using these samples, the average sensor values for each point log ($\bar{S}$), the average initial sensor values without contact ($S_{0}$), and the standard deviation from the average sensor value without contact ($\sigma_{0}$) were identified and used to calculate the signal-to-noise ratio (SNR) of each sensor using Eq. \ref{eq:SNR}. 
\begin{equation}
    SNR_i = 20\text{log}_{10}\left(\frac{max(\bar{S}_i) - S_{i,0}}{\sigma_{i,0}}\right)
    \label{eq:SNR}
\end{equation}
where $i$ is the sensor number.

\begin{figure}\vspace{+5pt}
    \includegraphics[width=0.48\textwidth]{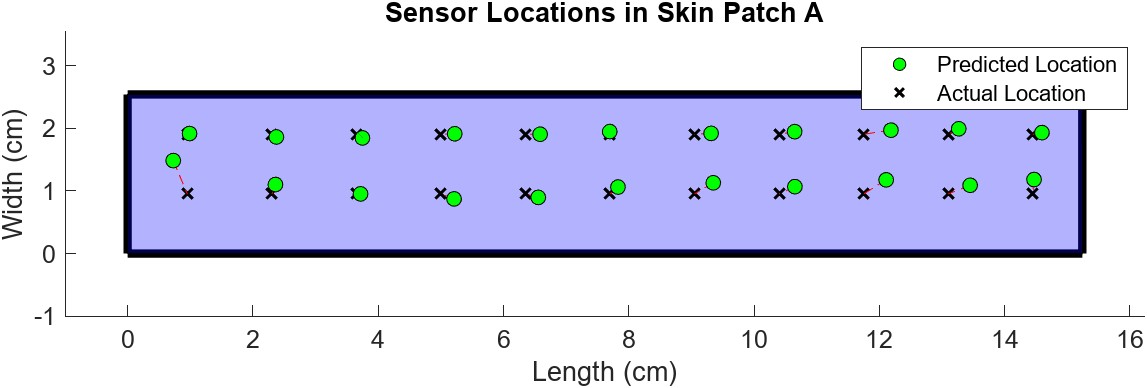}
    \includegraphics[width=0.48\textwidth]{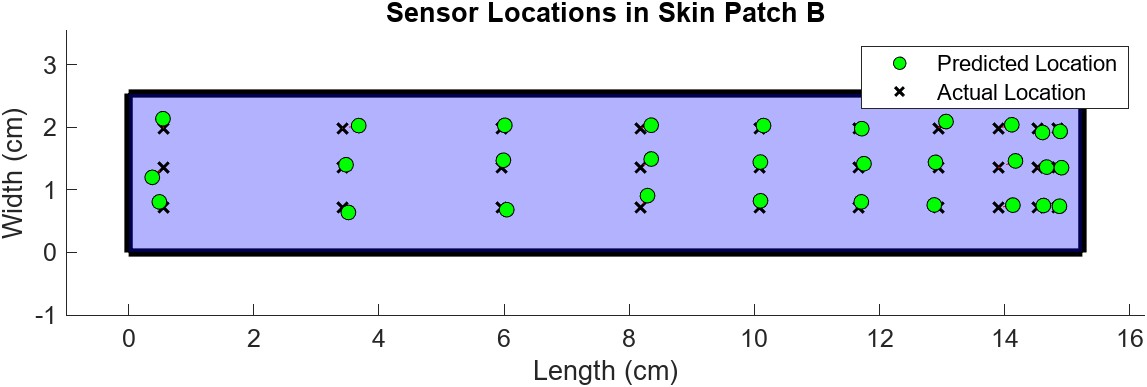}
  \caption{Final sensor location predictions using a uniform 5x20 grid of \textsl{point logs} on the surface of skin patch B. We can see that our localization method can determine both evenly and unevenly distributed sensor locations. The drift values associated with specific sensors can be attributed to errors in the positioning of the probe, data processing, and the number of point logs.}
  \label{fig:predictions}
\end{figure}

The distances between the predicted sensor locations and the true sensor locations were used to determine the accuracy of the localization method presented. Fig. \ref{fig:predictions} shows a visual representation of the prediction locations on the surface of \textsl{Patch B}.

\section{Results and Discussion}

The average SNR of all the sensors in \textsl{Patch A} and \textsl{Patch B} were found to be $61.3dB$ and $64.7dB$ respectively. The measurements made with the \textsl{VARSkin} were found to be reliable and consistent for each sensor in this test setup as a result of the high SNR values. Additionally, the SNR values found are on the same order of magnitude as findings by \cite{teyssier2021human} with a hardware gain of $1$ on a comparable artificial skin. Table \ref{tab:resultsB} shows SNR values and prediction results for a subset of sensors in \textsl{Patch B}, and Table \ref{tab:error} shows the standard deviation of all the location predictions in each skin patch. The standard deviation of \textsl{Patch A} and \textsl{B} were $2.6mm$ and $1.6mm$ respectively. The prediction results of \textsl{Patch A} were found to be larger than that of \textsl{Patch B} when measuring greater than 50 \textsl{point logs}.

A few sources of error are identified to affect the results in the localization method and are as follows:

\begin{figure}
  \centering
  \includegraphics[width=\linewidth]{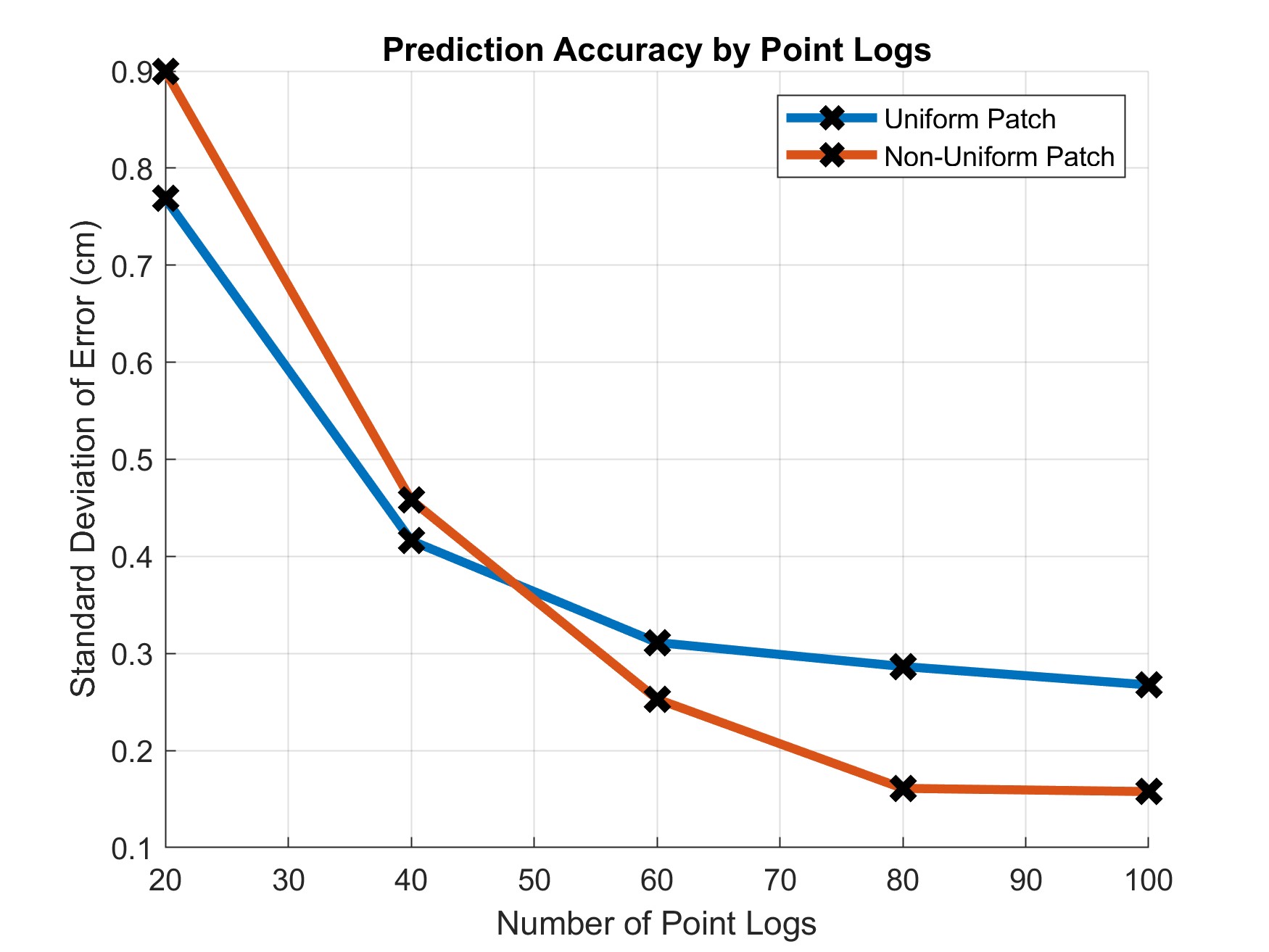}
  \caption{Standard deviation prediction error for varying quantities of \textsl{point logs} taken in a uniform grid. The prediction error inversely decreases with higher quantities of \textsl{point logs}.}
  \label{fig:accuracyBySamples}
\end{figure}

\subsubsection{Probe Error}
    The grid overlay projected on the skin patches for \textsl{point log} touch positions is accurate to the pixel resolution of the webcam used. The grid is generously estimated to have an error of $\leq 0.2mm$ using the width of the skin patch ($152.4mm$) divided by the pixel width resolution ($1280$). However, a larger portion of the error come from human probe positioning within the grid, as the surface contact area and exact location of the probe cannot be identified from a top-down observer's point of view to an accuracy greater than $\pm 2mm$.
    Although we experimented with multiple probe alignment methods, such as automatic probe alignment of a grounded metal probe set to always touch the skin surface with a specified contact force, they led to only marginal accuracy improvements at the cost of a slowing down of the data collection. 

\begin{table}
    \centering
    \begin{tabular}{|c|c|c|c|c|}
        \hline
        Sensor & Real & Predicted & Prediction & SNR \\ 
        Number & Position ($cm$) & Position ($cm$) & Error ($cm$) & (dB) \\
        \hline
        Sensor 1 & (0.55, 1.98) & (0.55, 2.13) & 0.15 & 68.5 \\ 
        Sensor 2 & (0.55, 1.35) & (0.38, 1.20) & 0.23 & 62.4 \\ 
        Sensor 3 & (0.55, 0.72) & (0.50, 0.81) & 0.11 & 67.9 \\ 
        ... & ... & ... &... & ... \\ 
        Sensor 28 & (14.84, 1.98) & (14.90, 1.93) & 0.07 & 67.3 \\ 
        Sensor 29 & (14.84, 1.35) & (14.92, 1.35) & 0.07 & 67.5 \\ 
        Sensor 30 & (14.84, 0.72) & (14.88, 0.74) & 0.04 & 69.4 \\ 
        \hline 
    \end{tabular}
    \caption{Prediction results and SNR values for a subset of sensors in \textsl{Patch B}. Prediction and SNR results for all sensors contributed to the final average prediction error and SNR values. }
    \label{tab:resultsB}
\end{table}

\begin{table}
    \centering
    \begin{tabular}{|c|c|c|}
        \hline
        Skin &  $\sigma_{PE}$ & Average\\
        Patch &  ($mm$) & SNR (dB)\\
        \hline
        A & $\pm 2.6$ & 61.3 \\
        \hline
        B & $\pm 1.6$ & 64.7\\
        \hline
    \end{tabular}
    \caption{Standard deviation of the prediction location $\sigma_{PE}$ and average signal-to-noise ratio from each sensor in each patch. The signal-to-noise ratios calculated indicate reliable signal ranges for use in the localization method presented.}
    \label{tab:error}
\end{table}

    \subsubsection{Data Processing Error} 
\begin{figure}
  \centering
  \includegraphics[width=\linewidth]{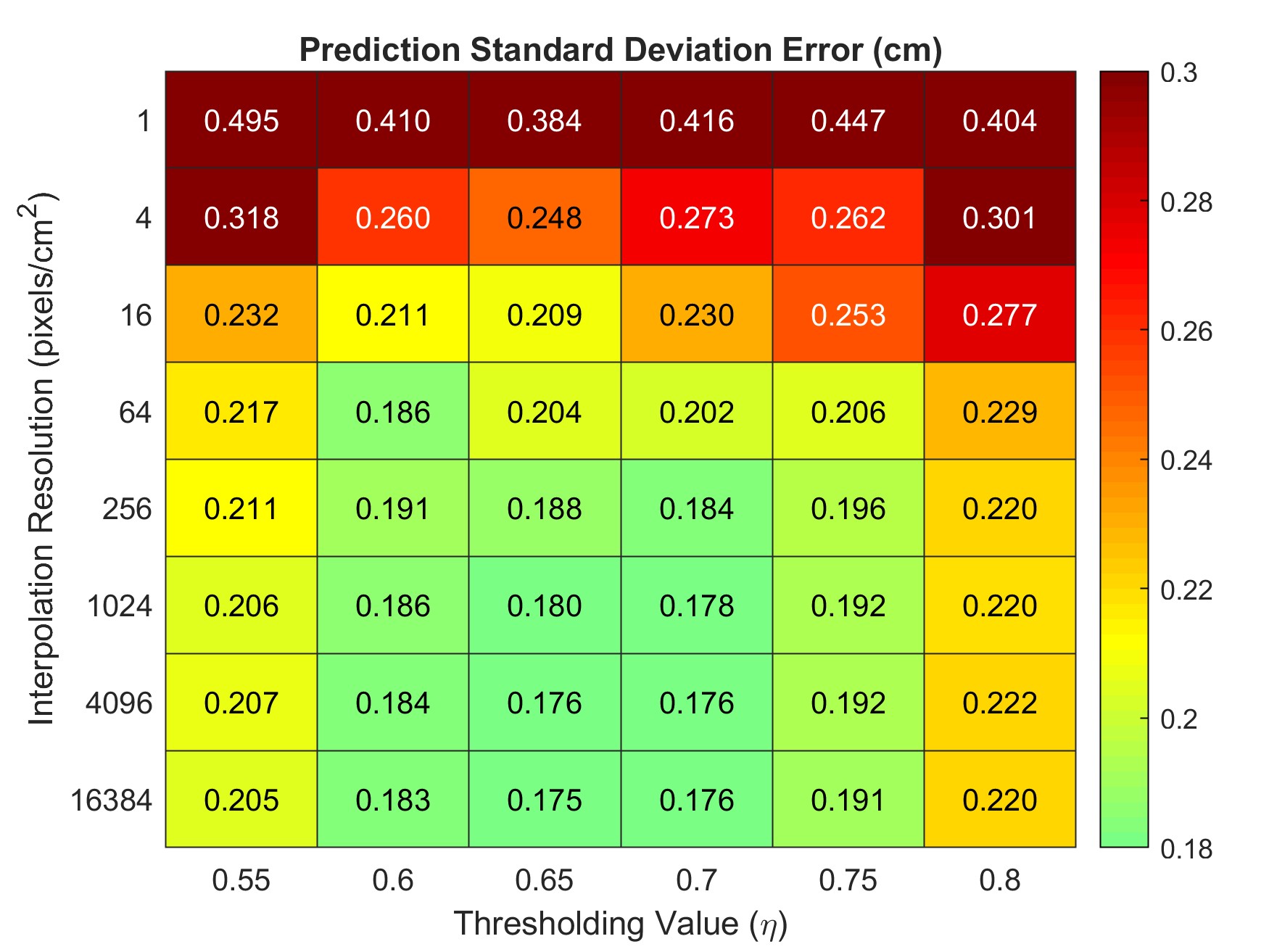}
  \caption{ Standard deviation of prediction error for varying threshold parameter $\eta$ and interpolation resolution. We see that
  to minimize the prediction error, we need to increase the interpolation resolution while keeping $\eta = 0.65 \pm 0.07$.}
  \label{fig:Errormap}
\end{figure}

    The interpolation resolution and threshold acceptance value are adjustable parameters in our upscaling and filtering processes. As shown in Fig. \ref{fig:Errormap}, higher interpolation resolution reduced prediction error, with smaller magnitudes of gained accuracy for exponentially increasing resolution. However, this comes at the cost of increasing computation time. Variation of the threshold displayed a consistent local minimum of prediction error in the general range of $\eta = 0.65 \pm 0.07$ for any given upscaling resolution.


    \subsubsection{Number of Point Logs}
    The number of \textsl{point logs} for each data set directly influences prediction accuracy. To test this parameter, \textsl{Patch A and B} sensor predictions were made using a varying \textsl{point log} quantity, $\eta = 0.65$ filtering threshold, 128 \textsl{pixel-per-cm} interpolation resolution, and a \textsl{point log} grid plan of uniform locations. An inverse relationship between prediction error and the number of \textsl{point logs} is observed in Fig. \ref{fig:accuracyBySamples}. The prediction results of \textsl{Patch A} were found to be larger than that of \textsl{Patch B} when measuring greater than 50 \textsl{point logs} as a result of the accumulation of the error sources listed above.


\section{Conclusion and Future Work}

In this work, we introduce \textsl{VARSkin}, a novel non-uniform mutual capacitance-based robot skin with variable sensor density. Additionally, we introduce a three-step sensor localization method that predicts the placement of all 30 intersection points in the non-uniform sensor array. Our \textsl{VARSkin} localization method is evaluated on two custom-built, flexible skin patches each with a concealed sensor layout. We observe a sub $2$mm positional accuracy for our mutual capacitance-based artificial skin. Through this work, we have shown that a variable density artificial skin with unknown sensor positions can be localized with common tools, lowering the barrier for entry for those who wish to create their own \textsl{VARSkin} sensors. This work could lead to easy-to-fabricate sensors that are large-scale, flexible, uniquely shaped, and variable in density.

The localization method is appropriate for artificial skin patches prior to mounting on curved surfaces and should be tested on complex three-dimensional configurations. Moreover, the sensor density limit of our localization method is unexplored and could allow for increased tactile data leading to finer interaction feedback. In future work, automatic probing complemented by real-time processing could allow for parallel calibration and execution with more efficient and accurate predictions. Future work in increasing the degree of accuracy could allow for the implementation of artificial skin platforms comprised of sensors on the scale of human skin densities and complexity.

\bibliographystyle{IEEEtran}
\bibliography{references}

\end{document}